\documentclass[pdflatex,sn-mathphys-num]{sn-jnl}
\usepackage{xstring}
\newcommand{\eqrefs}[1]{(\StrSubstitute{#1}{ }{,})}
\usepackage{graphicx}%
\usepackage{multirow}%
\usepackage{amsmath,amssymb,amsfonts}%
\usepackage{amsthm}%
\usepackage{mathrsfs}%
\usepackage[title]{appendix}%
\usepackage{xcolor}%
\usepackage{textcomp}%
\usepackage{manyfoot}%
\usepackage{booktabs}%
\usepackage{algorithm}%
\usepackage{algorithmicx}%
\usepackage{algpseudocode}%
\usepackage{listings}%
\usepackage{float}
\usepackage{hyperref}
\usepackage{cleveref}
\hypersetup{
	colorlinks,
	citecolor=black,
	filecolor=black,
	linkcolor=black,
	urlcolor=black
}

\begin{document}
%\tableofcontents

\title[Article Title]{Hybrid Deep Learning Model for epileptic seizure classification by using 1D-CNN with multi-head attention mechanism }

\author[1]{\fnm{Mohammed} \sur{Guhdar}}\email{mohammed.guhdar@uoz.edu.krd}

\author[1]{\fnm{Ramadhan J.} \sur{Mstafa}}\email{ramadhan.mstafa@uoz.edu.krd}

\author[1]{\fnm{Abdulhakeem O.} \sur{Mohammed}}\email{a.mohammed@uoz.edu.krd}

\affil[1]{\orgdiv{Computer Science Department}, \orgname{College of Science, University of Zakho}, \orgaddress{\country{Kurdistan Region, Iraq}}}

\abstract{
    An epileptic seizure is a neurological disorder that significantly impacts the lives of many individuals. According to the World Health Organization (WHO), approximately 10 out of every 1,000 people worldwide are affected by epilepsy. Developing an accurate prediction mechanism for seizures is crucial for effective management and treatment; however, this task remains challenging due to the complex, noisy, and non-stationary nature of electroencephalography (EEG) signals. This paper presents a hybrid deep learning model designed for the prediction and analysis of epileptic seizures with high accuracy. The model integrates wavelet transform, 1D convolutional layers, and multi-head attention mechanism, leveraging the strengths of convolutional neural networks (CNNs) in capturing spatial patterns while enabling simultaneous focus on various temporal aspects of the EEG signal. To enhance the model's robustness and generalization capabilities, we employed multiple regularization techniques, including dropout layers, L2 regularization, and early stopping, to effectively mitigate overfitting. Our model achieved a remarkable classification accuracy of 99.83\% on benchmark datasets, outperforming all existing models. These findings contribute significantly to the advancement of automated tools for epileptic seizure detection, paving the way for more accurate diagnoses and improved clinical application in prediction systems. 
}


\maketitle

\section{Introduction}
Epilepsy is a prevalent neurological disorder globally, impacting around 50 million people \cite{WHO_epilepsy_50million}. Epileptic seizures result from sudden abnormal electrical activity in the brain, which can be read as sudden and significant changes in the EEG signal of the brain. The signal can vary in severity and frequency, which results in loss of consciousness and muscle contractions for a short period of time \cite{epilepsyfoundation_myoclonic}. Individuals with epilepsy often face significant employment challenges due to safety concerns in certain work environments. Many jobs that involve working at heights, operating heavy machinery, or in other potentially hazardous settings may be restricted for people with seizure disorders. This certainly limits job options and economic opportunities for those living with epilepsy.

Recently this field has gained increasing attention, particularly due to the application of deep learning techniques and increasing number of people suffering from epileptic seizures around the world \cite{jeon2021increasing}. Recent advancements in deep learning, particularly the combination of Convolutional Neural Networks (CNNs) with attention mechanisms, have shown promising results in analyzing complex time-series data like EEG signals. The multi-head attention mechanism, initially proposed for natural language processing tasks, has proven effective in capturing long-range dependencies and focusing on relevant parts of the input sequence. 

Deep learning models, such as Convolutional Neural Networks (CNNs) and Long Short-Term Memory (LSTM) networks, have demonstrated significant potential in analyzing and classifying signals. Their adaptive learning and ability to extract features directly from raw data eliminate the need for manual feature extraction such as extracting features by hand, making them promising candidates for various time-series data classification and recognition tasks, such as seizure detection from EEG data as implemented by Roy et al.\cite{roy2019_RNN_4}. CNNs are excellent at extracting spatial features, while LSTMs are designed to capture temporal dependencies in sequential data \cite{zhu2020exploring_lstm_temporal_5}. The integration of these techniques has led to substantial performance enhancements in both spatial and temporal pattern recognition tasks \cite{raghu2019transfusion_temporalpattern_6}.Feature extraction methodologies dependent on manual engineering frequently prove insufficiency in capturing salient characteristics due to the inherent complexity and non-stationary nature of electroencephalographic signals. Thus, deep learning models have been quite suitable for the classification task of EEGs, as shown in \cite{ranjan2024deep_EEG_DEEP_7,dutta2024deep_EEG_DEEP_8,badr2024review_EEG_DEEP_9,dutta2024deep_EEG_DEEP_10}.

Previous studies have shown that even slight alterations in feature selection or scaling, can result in significant impacts on overall model performance. For instance, the application of feature selection techniques such as Principal Component Analysis (PCA) and Chi-square has been shown to improve model accuracy and performance by reducing noise and dimensionality \cite{rupapara2023chi_PCA_CHI_11,hussain2023feature_ChiSq_EEG_15}. \\Furthermore, these feature selection techniques have been used in prior research to improve the classification accuracy of EEG signals by identifying the most salient features \cite{hussain2023feature_ChiSq_EEG_15}.
However, it is important to note that integrating such techniques as PCA should be approached with caution. While these methods can enhance performance, any apparent improvements may occasionally be attributed to overfitting rather than genuine generalization. The efficacy of these preprocessing methods is inherently dependent on the deep learning architecture, the specific use case, and the overall complexity of the model. This interdependence underscores the necessity for careful consideration and rigorous evaluation when implementing these techniques in EEG signal analysis and epileptic seizure detection models.

Incorporating feature scaling and dimensionality reduction is essential in deep learning pipelines. Previous studies have demonstrated that scaling EEG data can enhance neural network performance \cite{chaudhari2023neural12}. Common techniques, such as standard scaling and MinMax scaling, have been widely adopted  \cite{de2023choice_MinMax13}. However, the comparative performance of different scaling techniques on specific model architectures, such as CNNs and LSTMs, remains inadequately explored. PCA is frequently employed to reduce the dimensionality of EEG data while preserving essential features \cite{blanco2024real_PCA_EEG_14}. Despite its use in improving model accuracy, determining the optimal percentage of variance to retain for peak performance continues to be an active area of research. 

Our research addresses the challenge of developing an enhanced deep learning architecture for epileptic seizure detection. We propose a novel hybrid model that integrates wavelet transform and Conv1D with an attention mechanism, aiming to improve seizure recognition accuracy. This innovative architecture not only achieves the highest accuracy among existing hybrid models but also highlights the importance of the attention mechanism in enhancing performance. By contributing to the refinement of automated seizure detection systems, our work has the potential to improve clinical decision-making in epilepsy management.\\

The main contribution of this study can be summarized as follow: 
\begin{itemize}
%\item Addresses the challenge of developing an enhanced deep learning architecture for epileptic seizure detection
\item A novel hybrid deep learning model is proposed that integrates wavelet transform with 1D convolutional layers (Conv1D) and a multi-head attention mechanism, specifically designed for epileptic seizure detection.
\item The proposed model achieves superior seizure recognition accuracy compared to existing hybrid models, demonstrating the efficacy of combining these techniques.
\item The study emphasizes the critical role of the attention mechanism in improving the model’s performance, particularly in capturing important temporal features in EEG data.
 \end{itemize}

\section{Related work}
Recently, the interest in applications of deep learning related to the field of
biomedical signal processing has garnered  a significant attention, particularly focusing on electroencephalogram (EEG), electrocardiogram (ECG or EKG), and electromyography (EMG) data. These physiological signals are complex by nature due to their non-stationary, susceptibility to noise, and sequential characteristics. Deep learning architectures have shown remarkable efficacy in analyzing and classifying such complex data modalities, especially when integrated with sophisticated preprocessing methodologies such as adaptive wavelet analysis, initially conceptualized by Erdol et al. \cite{erdol1996waveletAdaptive_wavelet_16}. Building upon this foundation, Ruiz et al. \cite{ruiz2024fully_16} introduced an algorithmic framework for extracting time-variant wave morphologies from non-stationary signals through harmonic component modeling, exhibiting superior performance in denoising, decomposition, and adaptive segmentation tasks.

The application of wavelet transform has emerged as a powerful technique for decomposing complex biomedical signals into multiple sub-signals of different levels, have been widely used for enhancing feature extraction in various data types under different scale. A notable example of this approach is the work of Zhao et al. \cite{zhao2020ecg_Wavelet_CNN_18}, who developed an innovative framework combining wavelet transform with a 24-layer Convolutional Neural Network (CNN). Their study, using the PhysioNet/CINC challenge dataset \cite{goldberger2000physiobank_dataset2000_19}, employed wavelet transform to decompose the original ECG signal into nine distinct sub-signals at various scales. This sophisticated decomposition enabled more refinement analysis, yielding promising results with an accuracy of 87.1\% and an F1 score of 86.46. These findings highlight the potential synergies between advanced signal processing techniques and deep learning architectures for ECG analysis, demonstrating how this integrated approach can improve the interpretation of complex data such as biomedical signals.

 Convolutional Neural Networks (CNNs) have gained increasing popularity across nearly all domains involving data. Their ability to extract hidden or meaningful patterns from complex data has made them excellent tool for tasks related to image classification, signal processing, natural language processing, and more. Li et al. \cite{li2022motor_CNN+LSTM} proposed a hybrid model that integrated CNN with the Long-short term memory (LSTM) network in parallel combination for motor imagery EEG classification on the BCI Competition IV dataset provided by Graz University \cite{brunner2008bci_dataset_17}. The model achieved an accuracy of 87.68\%, outperforming the traditional CNN or LSTM models alone. 

Support Vector Machines (SVMs) are a widely utilized classification algorithm recognized for their capacity in achieving high accuracy in classification related tasks, particularly when combined with suitable kernels.  Sadam et al.\cite{sadam2024epileptic_CNN+SVM_20}  combined SVM with CNN for seizure detection based on EEG data. The authors first transformed their data into scalogram images by using the continuous wavelet transform technique and then fed these images into their hybrid model named SD-CNN. The accuracy achieved by this hybrid model was around 94\%. It was quite interesting to transfer numerical data to image data and then feed them into the model; it surely affected performance negatively as EEG data are numerical by nature. 

Nahzat et al.  \cite{nahzat2021classification_PCA+21} conducted a comprehensive investigation into dimensionality reduction techniques for epileptic seizure classification, specifically focusing on principal component analysis (PCA). Their study utilized the widely-recognized UCI epilepsy dataset \cite{andrzejak2001indications_UCI_EEG_DATASET} and implemented multiple classification algorithms: artificial neural networks (ANN), K-nearest neighbors (KNN), random forests (RF), support vector machines (SVM), and decision trees (DT). The researchers systematically evaluated each algorithm's performance under two conditions:  with and without dimensionality reduction via PCA as a preprocessing step, to quantify the impact of dimensionality reduction on classification accuracy and computational efficiency. Their findings demonstrated that while the random forest algorithm achieved the highest accuracy of 97\% in both scenarios, the PCA-enhanced approach significantly improved computational performance while maintaining classification accuracy, suggesting the potential benefits of dimensionality reduction in epileptic seizure detection systems.

Tors et al. \cite{torse2021classification23} pointed out that manual observation of EEG readings is typically used to determine whether a person has epilepsy, which makes it challenging to develop an automated diagnostic system. Their study utilizes data from the Bonn EEG Dataset for analysis. They propose using Least Square Support Vector Machine (LSSVM), as it can handle linear equations more effectively. The accuracy achieved with this method was 94.7\%. 

Natu et al. \cite{natu2022retracted24} addressed challenges of noise removal from EEG signals. The integrity of the EEG signal often gets compromised due to extraneous background noises or involuntary movements of muscle tissues as they are recorded by EEG sensors, thereby complicating its automatic detection. Recognizing this limitation, the paper reviews various automated methodologies, noting that feature selection and classification concerning epilepsy are particularly laborious and error-prone. 

George et al. \cite{george2020epileptic25} proposed the ResNET-50 model as an automated system to define the classes of EEG data, They investigated the classification of interictal, preictal, and ictal phases of epileptic seizures using EEG data from the CHB-MIT, Freiburg, BONN, and BERN datasets. By transforming 1D EEG signals into 2D EEG images using time-frequency analysis, the proposed model achieved an accuracy of 94.88\%.

Hilal et al. \cite{hilal2022intelligent_26} proposed Deep Canonical Sparse Autoencoder-based Epileptic Seizure Detection and Classification (DCSAE-ESDC) framework integrates multiple techniques. Coyote Optimization Algorithm (COA) was employed for feature selection, while Deep Canonical Sparse Autoencoder (DCSAE) was used for classification and detection. To further optimize model performance, Krill Herd Algorithm (KHA) was utilized for parameter tuning . The model was applied to the epileptic seizure dataset \cite{andrzejak2001indications_UCI_EEG_DATASET} and achieved an accuracy  of 98.67\%.

In a study by Liu et al. \cite{liu2023eeg_attentionMechanism} demonstrated the pivotal role of attention mechanisms in EEG-based emotion recognition. By focusing on salient EEG features, attention mechanisms enable models to effectively capture the most relevant information, a principle that extends to other biosignals like ECG and EMG. Building upon this insight, Liu et al. \cite{liu2023eegAP-CapsNet} introduced the AP-CapsNet model, where an attention mechanism is strategically integrated to capture global EEG features. This innovative approach enhances the network's ability to discern intricate relationships between EEG channels, leading to more accurate emotion recognition. The proposed model's efficacy is attributed to its capacity to extract the relative distribution of EEG characteristics through the combined power of attention mechanisms and pre-trained convolutional capsule networks. This synergy significantly elevates the performance of emotion detection and classification tasks.

While the studies presented demonstrate progress in applying deep learning to biomedical signal processing, several weaknesses are apparent. Many of the reported accuracies, such as 87.1\% by Zhao et al. \cite{zhao2020ecg_Wavelet_CNN_18}, 87.68\% by Li et al. \cite{li2022motor_CNN+LSTM}, and 94.7\% by Tors et al. \cite{torse2021classification23}, fall below 95\%, which can be considered risky for medical applications where high precision is crucial. The approach by Sadam et al. \cite{sadam2024epileptic_CNN+SVM_20} and George et al. \cite{george2020epileptic25} , which converts signals to scalogram images before processing, potentially increases time complexity without demonstrating clear superiority over direct numerical analysis. Additionally, the reliance on specific datasets, such as the UCI dataset or the Bonn EEG Dataset, may limit the generalizability of the findings to diverse patient populations or real-world clinical settings. The study by Natu et al. \cite{natu2022retracted24} highlights ongoing challenges in noise removal from EEG signals, indicating that this fundamental issue still impacts the field's progress. Furthermore, the variety of methodologies employed across these studies, from CNNs and SVMs to more complex hybrid models, suggests a lack of consensus on the most effective approach for biomedical signal analysis. 

\section{Materials and Methods}

\subsection{Dataset}

This study utilizes the UCI Epileptic Seizure Dataset \cite{andrzejak2001indications_UCI_EEG_DATASET} for the training and evaluation of our proposed model. This dataset serves as a comprehensive resource for epilepsy research, encompassing a diverse range of EEG recordings from multiple subjects.
The dataset is structured into five distinct epilepsy classes, each represented by 100 individual files, yielding a total of 500 subjects. Each file contains a 23.6-second recording of brain activity captured via EEG, sampled at a high temporal resolution of 4097 data points. Consequently, the dataset provides a robust foundation for analysis, offering 4097 data points for each of the 500 subjects.

\subsubsection{Data preprocessing}
   The 4097 data points for each subject are divided into 23 segments, each segment contains 178 data points for 1 second of EEG recording, final resulted dataset consist of 11,500 labeled samples (23 segments × 500 subjects), with total of 178 + 1  (y- output) column: $y \in \{ 1,2,3,4,5 \}$.

\begin{figure}
    \centering
    \includegraphics[width=0.9\textwidth]{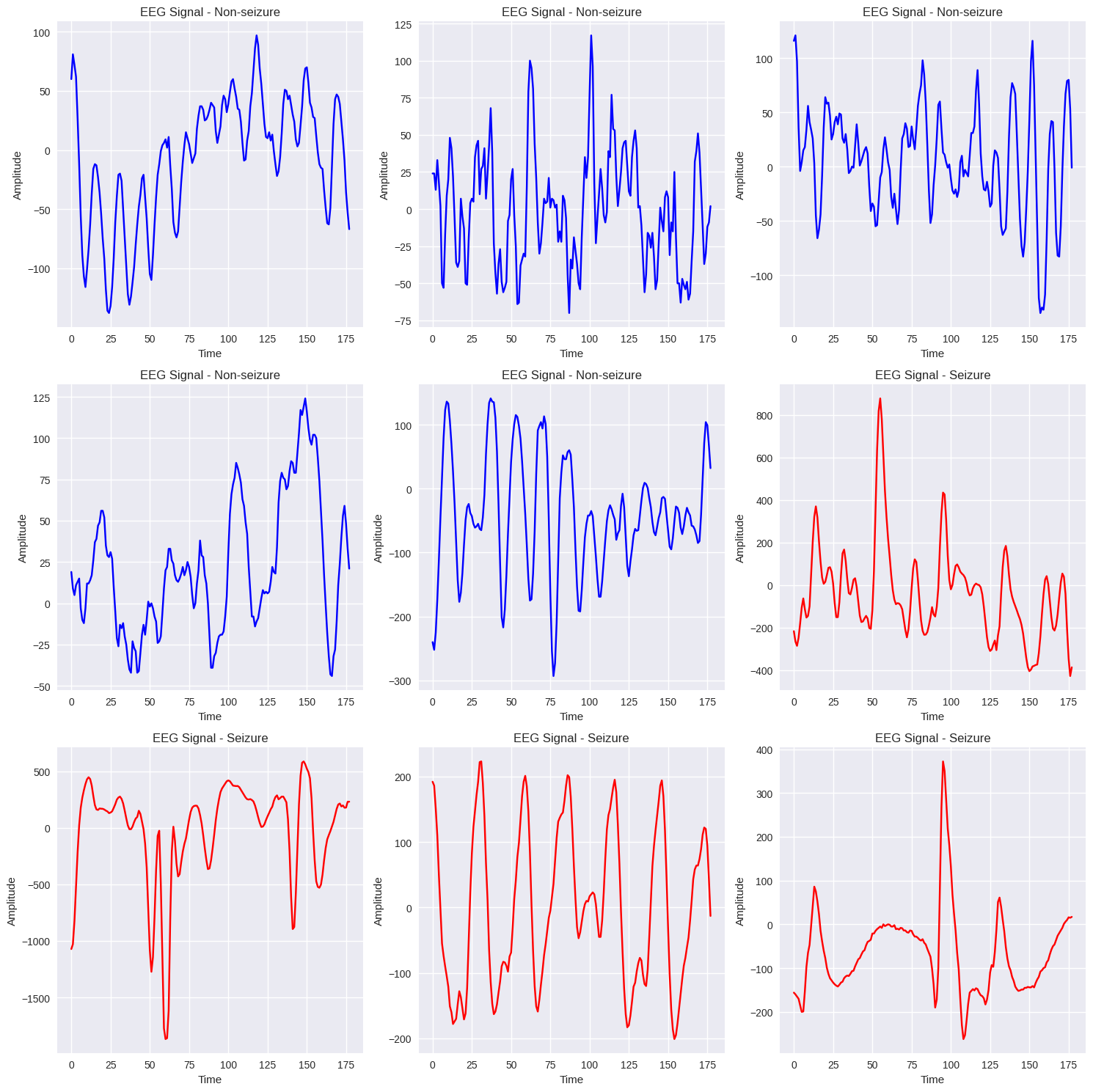}
    \caption{EEG Signal Patterns During Seizure (Red) and Non-Seizure (Blue) States}
    \label{fig:signal_shape}
\end{figure}    
\subsubsection{Label descriptions and adjustments}
The dataset is labeled as follows:

\begin{enumerate}
\item Epileptic seizure: EEG recordings during active seizure events.
\item Tumor area: EEG recordings from the region where a tumor is located, but not during a seizure.
\item Healthy brain, tumor present: EEG recordings from healthy brain areas in patients with a confirmed tumor in another region.
\item Eyes closed: EEG recordings from healthy subjects with their eyes closed.
\item Eyes open: EEG recordings from healthy subjects with their eyes open.
\end{enumerate}

In the context of epileptic seizure detection, category 1 represents the positive class (seizure), while categories 2–5 collectively form the negative class (non-seizure). This diverse set of non-seizure recordings allows the model to distinguish seizure activity not only from normal brain activity but also from other abnormal brain states and different levels of consciousness.

Given the focus of this research on epileptic seizure detection and classification, we have restructured the dataset to facilitate a binary classification task. The original labels were modified as follows: instances labeled "1" were retained to represent seizure events, while labels 2 through 5 were consolidated and reassigned a label of "0" to denote non-seizure instances. This reconfiguration simplifies the classification task to a binary problem of seizure versus non-seizure detection. Figure \ref{fig:signal_shape} illustrates representative samples of both seizure and non-seizure signals after this label adjustment.

\subsection{Feature Scaling}

Prior to model training, feature scaling was performed on the EEG data using StandardScaler to normalize the features. This preprocessing step is crucial for EEG signal analysis as it ensures all features contribute equally to the model and helps mitigate the impact of varying amplitudes across different EEG recordings. StandardScaler transforms the data by centering it to have zero mean and unit variance, following the Eq. \eqref{eq1} \cite{thara2019auto_featurescalingequation}:

\begin{equation}
    z = \frac{x - \mu}{\sigma}   \label{eq1}
\end{equation}

where $z$ is the normalized value, $x$ is the original feature value, $\mu$ is the mean of the feature values, and $\sigma$ is the standard deviation. This standardization is particularly important for our deep learning model as it helps accelerate convergence during training and ensures that features with larger magnitudes do not dominate the learning process. Additionally, standardization makes the training less sensitive to the scale of features, which is essential when dealing with EEG signals that may have different scales across channels or recordings.

\subsection{Performance Evaluation Metrics}

To rigorously evaluate the performance of the proposed model, we employed a diverse set of evaluation metrics. The main metrics utilized were accuracy, precision, recall, F1 score, Critical Success Index (CSI), and Matthews Correlation Coefficient (MCC) as in Eqs. \eqrefs{\ref{eq2} \ref{eq3} \ref{eq4} \ref{eq5} \ref{eq6} \ref{eq7}} respectively \cite{powers2007evaluation_Pression,powers2020evaluation_F1,mbizvo2023using_CSI,chicco2023matthews_MCC}, defined mathematically as follows:

\begin{equation}
\text{Accuracy} = \frac{\text{TP} + \text{TN}}{\text{TP} + \text{TN} + \text{FP} + \text{FN}} \label{eq2} \qquad 
\end{equation}

\begin{equation}
\text{Precision} = \frac{\text{TP}}{\text{TP} + \text{FP}} \label{eq3}
\end{equation}

\begin{equation}
\text{Recall} = \frac{\text{TP}}{\text{TP} + \text{FN}} \label{eq4}
\end{equation}

\begin{equation}
\text{F1 Score} = \frac{2 \times (\text{Precision} \times \text{Recall})}{\text{Precision} + \text{Recall}} \label{eq5}
\end{equation}

\begin{equation}
\text{CSI} = \frac{\text{TP}}{\text{TP} + \text{FN} + \text{FP}} \label{eq6}
\end{equation}

\begin{equation}
%\scriptsize
\small
\text{MCC} = \frac{\text{TP} \times \text{TN} - \text{FP} \times \text{FN}}{\sqrt{(\text{TP} + \text{FP})(\text{TP} + \text{FN})(\text{TN} + \text{FP})(\text{TN} + \text{FN})}} \label{eq7}
\end{equation}

These diverse set of metrics provides unique insights into the model's performance: accuracy quantifies the overall correctly predicted instances, precision measures the positive predictive value, and recall indicates the model's ability to identify actual positive cases. The F1 score offers a balanced measure between precision and recall, particularly useful for imbalanced datasets. CSI (Critical Success Index), which disregards true negatives, is especially pertinent for rare event prediction, while MCC (Matthews Correlation Coefficient) provides a balanced evaluation even with disparate class sizes, effectively summarizing the confusion matrix of binary classification problems.

\section{Proposed Method}

This study presents a novel deep learning model that integrates multiple techniques to optimize performance in epileptic seizure detection. The proposed architecture initiates with a wavelet transform preprocessing step, effectively mitigating noise in EEG signals. This preprocessing is followed by a series of three 1D Convolutional layers, characterized by progressively increasing filter sizes and decreasing kernel sizes. Each convolutional layer is augmented by batch normalization and MaxPooling operations to enhance feature extraction and reduce computational complexity.

A pivotal component of the model is the Multi-Head Attention layer, comprising 4 heads with a key dimension of 32. This mechanism enables the model to focus on diverse aspects of the input sequence simultaneously, potentially capturing complex temporal dependencies in the EEG data.

To further enhance the model's learning capacity, the architecture incorporates a skip connection. This technique combines the output of the final convolutional layer with earlier layers, facilitating the training of deep networks by preserving information from initial layers throughout the learning process.

\begin{figure}
    \centering
    \includegraphics[width=1\textwidth]{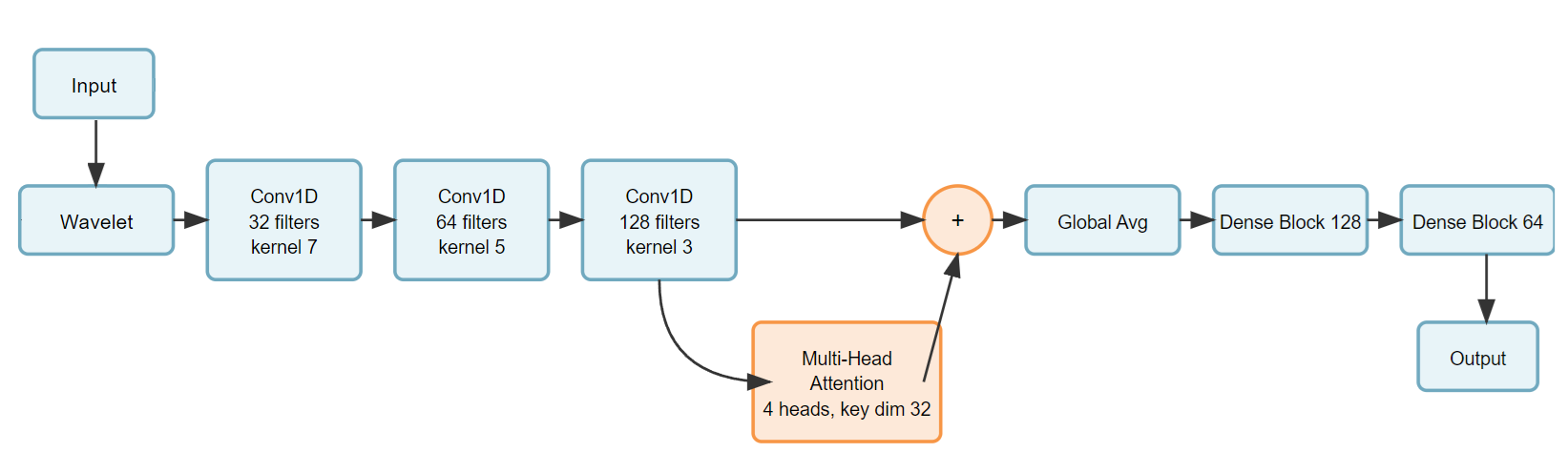}
    \caption{Proposed hybrid deep learning architecture for epileptic seizure detection}
    \label{fig:enter-label}
    \centering
\end{figure}

The proposed model follows a conventional training and testing methodology criteria, as illustrated in Figure \ref{fig:enter-label}. The process starts with the UCI Epileptic Seizure Dataset preporcessing, which inherently contains five output labels (1 to 5). The dataset subsequently undergoes preprocessing steps, including label adjustment, data cleaning, normalization, and feature scaling, to prepare it for model input. Following preprocessing, The dataset is divided into testing and training sets in order to assess the performance of the model.

 \subsection{Model Implementation}
 The proposed architecture, with a combination of 1D CNN and an attention mechanism as its main components, consists of three convolutional layers followed by an attention mechanism. Here, the input shape of the model is represented as (178, 1), which is 178 data points from the EEG signal. The Conv1D layer consists of 32 filters with a kernel size of 7. Then, the second layer uses 64 filters with a kernel size of 5, followed by the final layer, which has 128 filters with a kernel size of 3. BatchNormalization and MaxPooling1D follow each convolutional layer.\\
The most salient component of this model is a MultiHeadAttention layer with 4 heads and a key dimension of 32, combined via a skip connection with the output of the last convolutional layer. This is followed by LayerNormalization and GlobalAveragePooling1D. This model is followed by two dense layers of neurons equal to 128 and 64, with BatchNormalization and dropout of 0.5.
The last layer consists of one neuron with a sigmoid activation function, which outputs binary classification for seizure instances. The Adam optimizer is utilized, while binary cross-entropy is set as the loss function. In order to overcome overfitting, an L2 regularization of 0.001 is imposed on both convolutional and dense layers.

\subsection{Single-Level Wavelet}
\begin{figure}[H]
    \centering
    \includegraphics[width=0.9\textwidth]{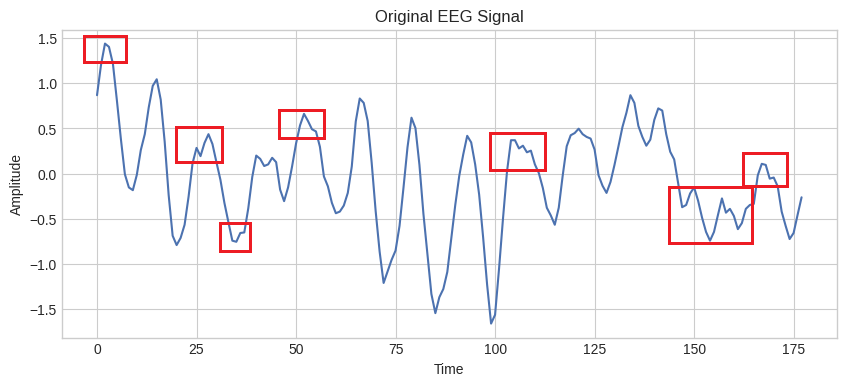}
    \caption{EEG signals before applying Wavelet Transformer}
    \label{fig:EEG_before_wavelet}
\end{figure}

\begin{figure}[H]
\centering
  \includegraphics[width=0.9\textwidth]{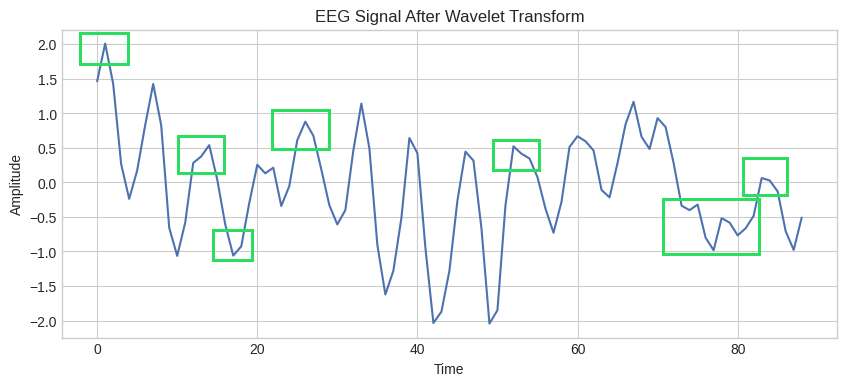}
    \caption{EEG signal after applying Wavelet transformer}
\label{fig:EEG_after_wavelet}

\end{figure}

A single-level wavelet transform utilizing the Daubechies 1 (db1) wavelet was applied to our EEG data to address noise in a minimalist manner. This approach was selected over more complex wavelets, such as Symlets or Coiflets, to preserve the essential characteristics of EEG signals. The db1 wavelet, being one of the simplest, facilitates subtle noise reduction while maintaining crucial signal features vital for epileptic seizure detection.
The impact of these minor alterations is illustrated in Figure \ref{fig:EEG_before_wavelet}, which depicts the signal prior to wavelet transformation, and Figure \ref{fig:EEG_after_wavelet}, which demonstrates how the signal becomes marginally smoother post-transformation. Notably, this process retains both high- and low-frequency data, which are critical components in EEG signals for accurate seizure detection.
These preprocessing steps enhance the efficiency of EEG pattern analysis, potentially improving the overall performance of the seizure detection model.

\subsection{1D Convolutional Neural Network (CNN)}
\begin{figure}
    \centering
    \includegraphics[width=0.9\textwidth]{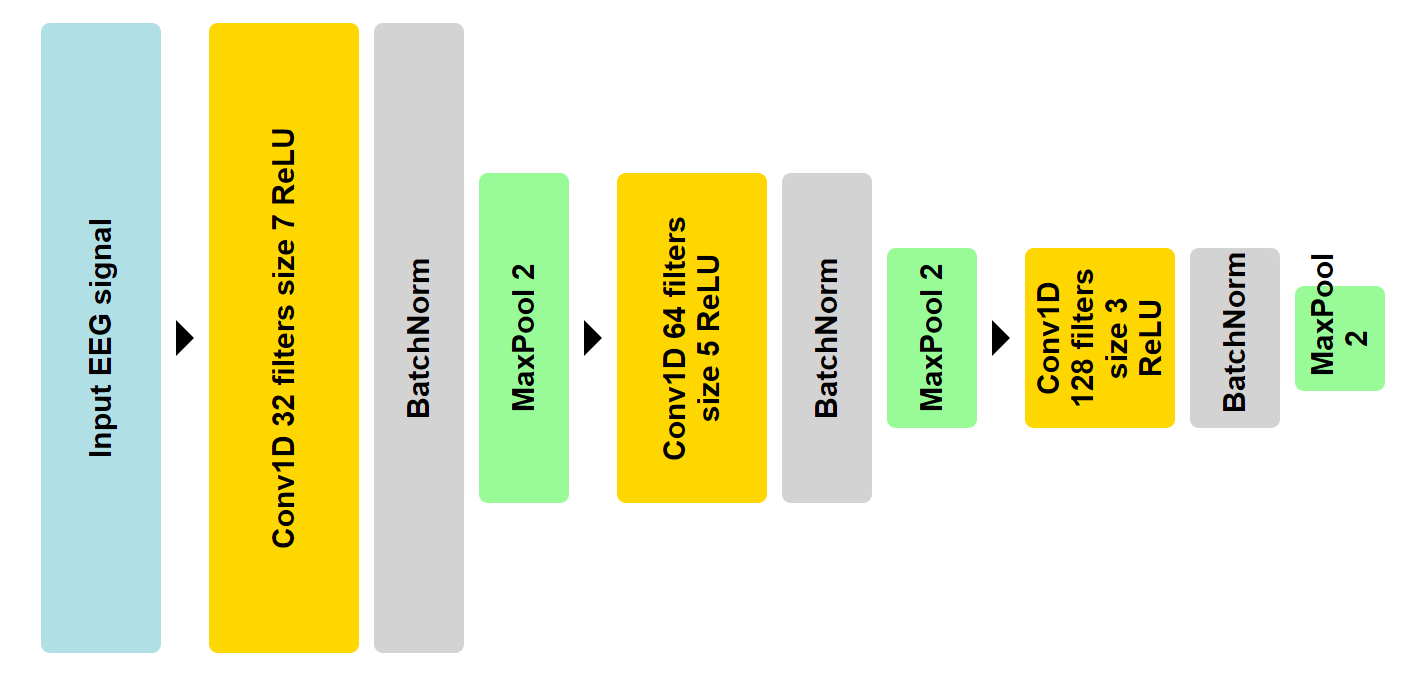}
    \caption{1D Convolutional Neural Network block of the proposed model.}
    \label{fig:1Dconv}
\end{figure}

Convolutional Neural Networks (CNNs) have demonstrated considerable promise in extracting spatial and temporal dependencies from EEG signals. The proposed model employs one-dimensional convolutional layers (Conv1D) to extract hierarchical features from electroencephalogram (EEG) signals. These 1D-convolutional layers incorporate filters of varying sizes (7, 5, and 3) alongside an increasing number of channels (32, 64, and 128) to recognize increasingly complex patterns within the EEG dataset. The structure of the 1D-convolutional model is illustrated in Figure \ref{fig:1Dconv}.
To enhance the model's generalization capability, L2 regularization is applied to both the convolutional and dense layers. This regularization technique helps mitigate overfitting by introducing a penalty term to the loss function based on the squared magnitude of the weights. Furthermore, batch normalization layers are implemented after each convolutional and dense layer. While not strictly a regularization technique, batch normalization can exert a regularizing effect by reducing internal covariate shift, thus potentially improving the model's performance and stability.

\subsection{Regularization and handling Overfitting}
\begin{figure}[H]
    \centering
    \includegraphics[width=0.9\textwidth]{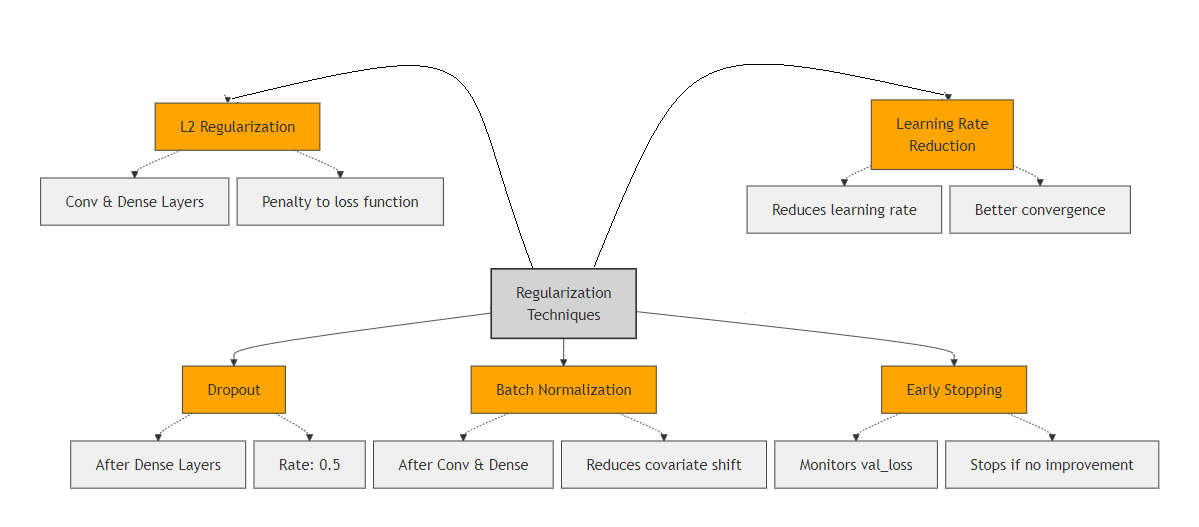}
    \caption{Regularization techniques were applied in the proposed model to prevent overfitting and improve performance.}
    \label{fig:regularization}
\end{figure}

The proposed model highly concentrated on overfitting problem, and to prevent overfitting several techniques has been applied as shown in Figure \ref{fig:regularization} which are: 
\begin{enumerate}
    \item L2 Regularization: This is applied to the convolutional and dense layers using the kernel\_regularizer=L2(0.001) parameter. L2 regularization mitigates overfitting by incorporating a penalty term into the loss function, which is contingent upon the squared size of the weights.
    \item Dropout: At a rate of 0.5, dropout layers are introduced subsequent to the dense layers. To reduce overfitting, dropout randomly deactivates a fraction of input units to 0 throughout the training process.
    \item Batch Normalization: BatchNormalization layers are used after each convolutional layer and dense layer. While not strictly a regularization technique, batch normalization can have a regularizing effect by reducing internal covariate shift.
    \item Early Stopping: The EarlyStopping callback is used to monitor the validation loss and halt training if it doesn't improve for a specified amount of epochs. By halting training when the model begins to overfit the training set, this helps avoid overfitting.
    \item  Reduction of Learning Rate: When the validation loss ceases to improve, the learning rate is decreased using the ReduceLROnPlateau callback. By doing this, overfitting can be prevented and the model can converge to a better minimum.
    
\end{enumerate}

These techniques work together to help regularize the model and prevent overfitting.  

\subsection{Multi-Head Attention Mechanism}
\begin{figure}
    \centering
    \includegraphics[width=0.9\textwidth]{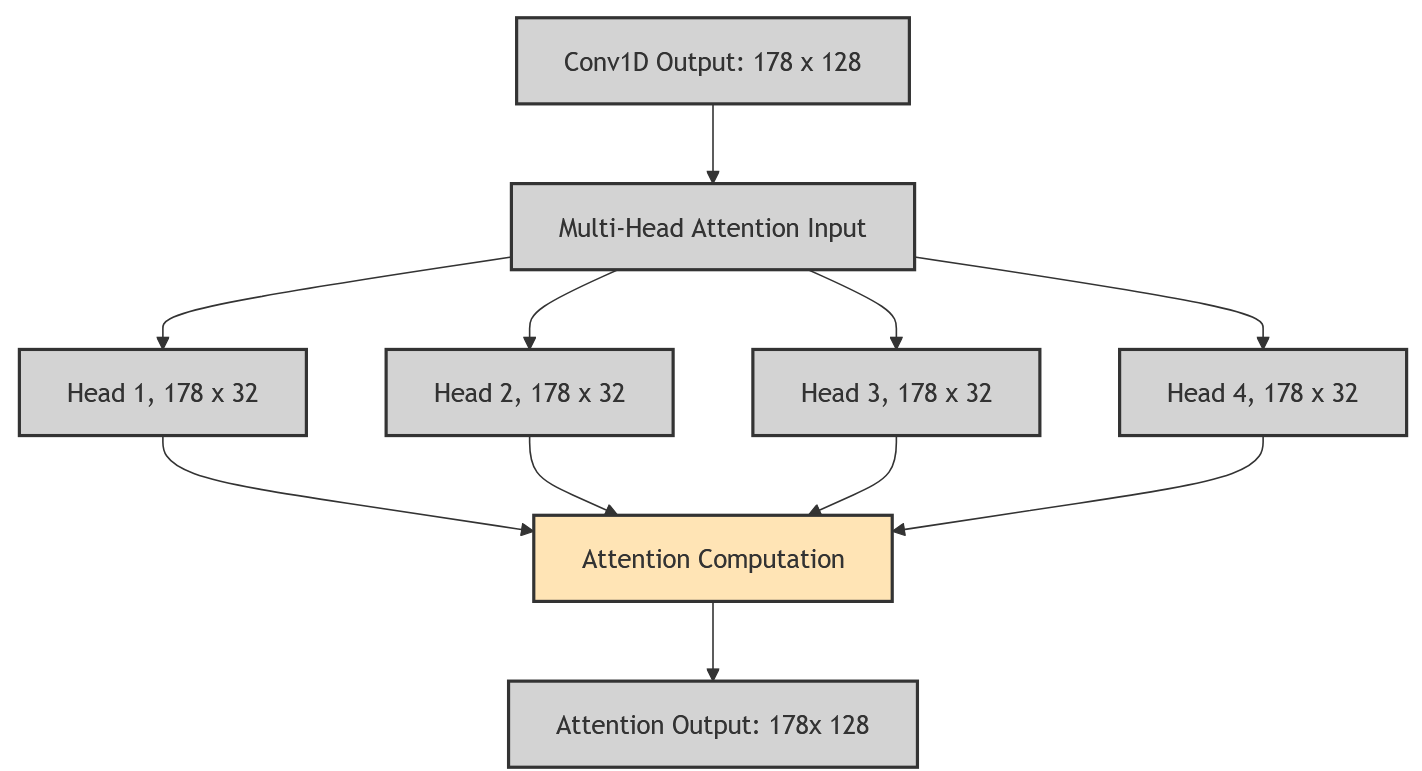}
    \caption{Multi-head attention mechanism with four heads}
    \label{fig:Multi-head attention}
\end{figure}

\begin{figure}
\centering
  \includegraphics[width=0.9\textwidth]{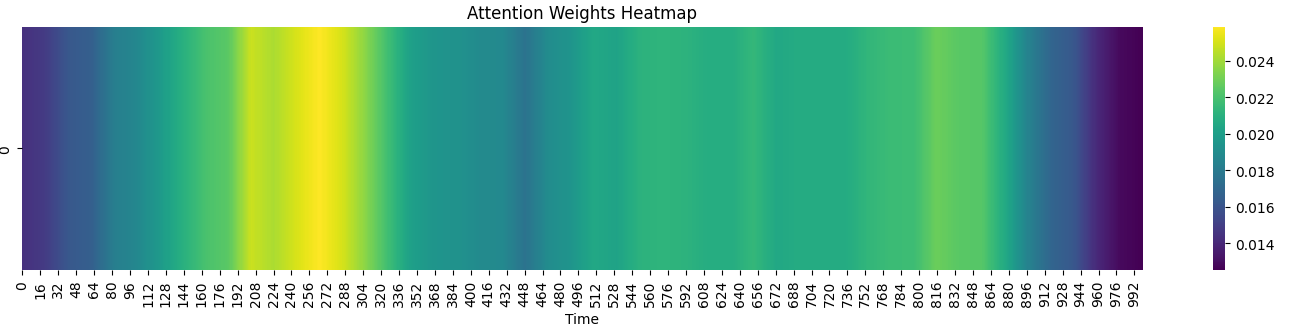}
    \caption{Multi-head attention mechanism heat-map for weighting process}
\label{fig:heat_map_weight}

\end{figure}

The most salient element of this architecture for extracting complex patterns within the EEG is an attention layer combination with 1D CNN. A Multi-Head Attention mechanism with 4 heads a key dimension of 32, allows the model to focus on various segments of the EEG sequence simultaneously. This approach leads to improved success in capturing subtle patterns indicative of an epileptic seizure.
Furthermore, this research incorporates a skip connection that integrates the output of the last convolutional layer (with a filter size of 128) with the output produced by the attention layer. The skip connection, also known as a residual connection, serves to bypass the attention layer. Such an approach has proven beneficial for training deep networks as it preserves information from earlier layers and mitigates issues related to vanishing gradients. The synergy between the attention mechanism and residual connections potentially enables our model to capture both complex and broad patterns in EEG signals, thereby enhancing seizure detection capabilities. Figure \ref{fig:Multi-head attention} illustrates the details of the 4-headed attention mechanism used in our model, while Figure \ref{fig:heat_map_weight} demonstrates how the attention mechanism operates by assigning weights to each value in the signal segment.

The integration of these advanced techniques—1D CNNs, multi-head attention, and skip connections—creates a robust architecture capable of capturing multi-scale temporal dependencies in EEG signals. The 1D CNNs extract local features at various scales, While the attention mechanism enables the model to concentrate on the most crucial part of the input sequence. The skip connection ensures that both fine-grained details from earlier layers and high-level features from later layers contribute to the final prediction. This comprehensive approach potentially enables the model to detect subtle EEG variations that may be indicative of an impending seizure, even in complex or noisy data. The synergistic effect of these components may provide our model with enhanced discriminative power, potentially leading to more accurate and reliable epileptic seizure detection.

\subsection{Global Average Pooling}
After the attention layer, we apply global average pooling to reduce the spatial dimensions of our feature maps. This helps in avoiding overfitting by reducing the total parameters in the model while still preserving the majority of global features that are important. For EEG signals, global average pooling will be useful in preserving the overall signal characteristics, which are crucial for seizure identification. 

\subsection{Fully Connected Layers with Dropout}
The last part of our model consists of two fully connected- dense layers with 128 and 64 units, respectively. In this way, the model learns higher-order representations of the EEG data. To reduce overfitting—a common problem in deep learning models—we apply Dropout with a rate of 0.5 between these layers. Adding L2 regularization with a factor of 0.001 reduces model over-fitting and improves generalization.

\section{Results and discussion}
This section presents an evaluation of our proposed model's performance and its comparison with existing state-of-the-art methods using the same dataset. We first detail our model's results across multiple evaluation metrics, including accuracy, Matthews Correlation Coefficient (MCC), Critical Success Index (CSI), and F1 score, supported by visual representations through accuracy/loss curves and a confusion matrix. The comparative analysis then contextualizes these results against other recent approaches, demonstrating how our proposed architecture advances the current state of epileptic seizure detection methodology.

\subsection{Proposed Method Performance}

\begin{figure}[H]
    \centering
    \includegraphics[width=0.9\textwidth]{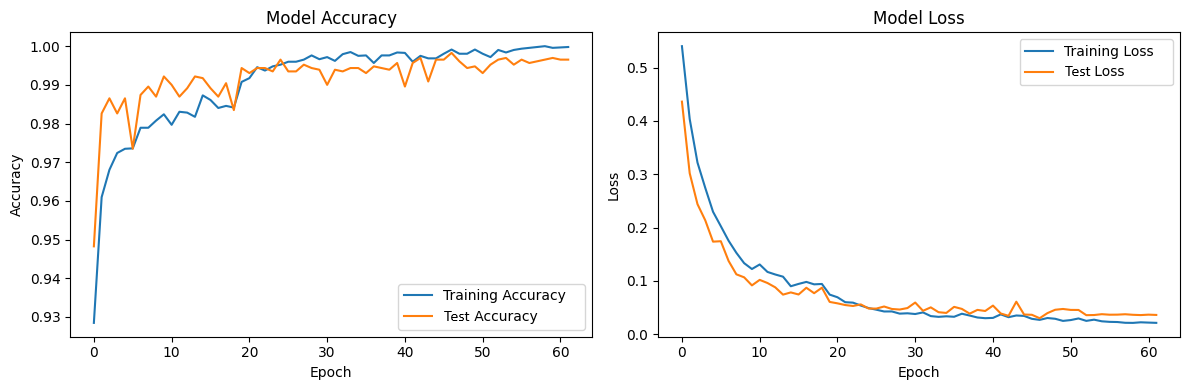}
    \caption{Model Accuracy and Loss}
    \label{fig:Accuracy and Loss}
    \centering
\end{figure}

To rigorously evaluate our proposed method, we implemented a standard data split protocol, allocating 80\% of the dataset for training and reserving the remaining 20\% for testing. The model's performance is illustrated in Figure \ref{fig:Accuracy and Loss}, which demonstrates the progression of both training and validation metrics over epochs.

\begin{figure}[H]
    \centering
    \includegraphics[width=0.9\textwidth]{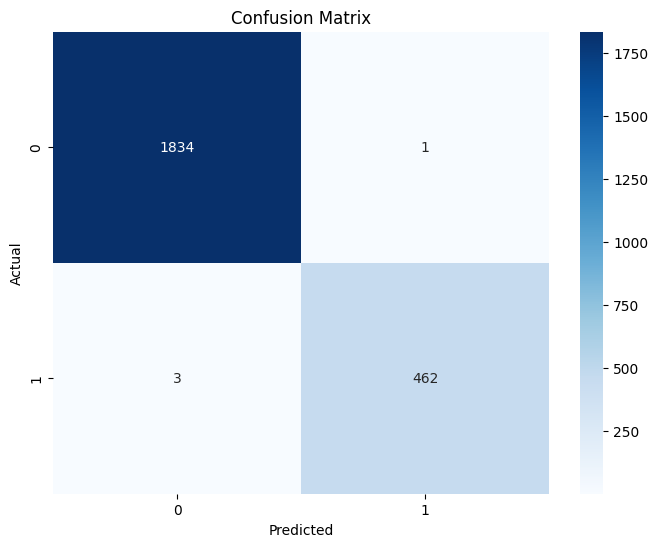}
    \caption{Confusion matrix of the proposed model}
    \label{fig:Confusion matrix}
    \centering
\end{figure}

The confusion matrix, presented in Figure \ref{fig:Confusion matrix}, provides a detailed breakdown of the model's classification performance on the test set. Out of 2300 test samples, the model correctly identified 1834 true negatives and 462 true positives, with only 4 miss-classifications (1 false positive and 3 false negatives). This result translates to an impressive overall test accuracy of 99.83\%, with a slight tendency towards under predicting positive cases.

Further performance evaluation metrics, presented in Table \ref{tab:performance_metrics}, offer a comprehensive view of the model's capabilities across multiple assessment criteria. These results collectively underscore the robustness and efficacy of our proposed approach in accurately detecting epileptic seizures from EEG data with high values across multiple assessment criteria.

\begin{table}[h]
\centering
\caption{Performance Metrics of proposed model for Seizure Detection }
\label{tab:performance_metrics}
\begin{tabular}{lc}
\hline
\textbf{Metric} & \textbf{Value} \\
\hline
Training Accuracy (\%) & 99.97 \\
Test Accuracy(\%) & 99.83 \\
Matthews Correlation Coefficient (MCC) & 0.9950 \\
Critical Success Index (CSI) & 0.9990 \\
F1 Score & 0.9990 \\
\hline
\end{tabular}
\end{table}

\begin{table}[h]
\caption{Comparison of Models Applied to the UCI Epilepsy Seizure Dataset}\label{tab:model_comparison}
\begin{tabular}{@{}lll@{}}
\toprule
\textbf{Model}  & \textbf{Accuracy} & \textbf{Publication Year} \\
\midrule
Long-Short Term Memory (LSTM) \cite{kunekar2024detection_LSTM97} & 97.1\% & 2024 \\
Conv1D+LSTM \cite{omar2024optimizing_993_CONV+LSTM} & 99.3\% & 2023 \\
HyEpiSeiD - CNN+Gated Recurrent Network \cite{bhadra2024hyepiseid9901_HyEpiSeiD} & 99.01\% & 2024 \\
SCA-Optimized LightGBM Classifier \cite{abenna2022eegSCA99} & 99.00\% & 2022 \\
Random Forest (RF) \cite{almustafa2020classification9708_RandomForest} & 97.08\% & 2020 \\
Random Forest (RF) \cite{kunekar2024comparison977_RandomForest2} & 97.7\% & 2023 \\
K-Nearest Neighbors+PCA \cite{nahzat2021classification_KNNPCA} & 99\% & 2022 \\
Conv1D+LSTM+Bayesian \cite{jain2024bayesian_1Dconv+LSTM+Bayesian} & 99.47\% & 2024 \\
FUPTBSVM (hybrid model) \cite{gupta2024functional8866FUPTBSVM} & 88.66\% & 2023 \\
\textbf{Proposed model} & \textbf{99.83\%} & \textbf{2024} \\
\bottomrule
\end{tabular}
\end{table}

\subsection{Comparative analysis}
Table \ref{tab:model_comparison} presents a comprehensive comparison of various models for epileptic seizure detection that used the same dataset, showcasing the evolution and improvement of techniques and interest over recent years. The proposed model achieves the highest accuracy at 99.83\%, surpassing other state-of-the-art approaches. The closest competitors are the Conv1D+LSTM+Bayesian model, which has achieved 99.47\% accuracy [40], and the Conv1D+LSTM model, which reached 99.3\% accuracy [34]. These top-ranking models prove the efficiency of combining convolutional and recurrent neural network architectures for EEG signal analysis. The HyEpiSeiD model [35] and the SCA-Optimized LightGBM Classifier [36] demonstrate remarkable performance as well, achieving accuracies of 99.01\% and 99.00\%, respectively, highlighting the effectiveness of hybrid and optimized methodologies in machine learning.

\section{Conclusion}
Our hybrid deep learning model combines 1D convolution layers with a multi-head attention mechanism to achieve high accuracy of 99.83\%, surpassing other existing state-of-the-art models utilized the same dataset. This success indicates that our approach shows promising potential for improving the accuracy of automated seizure detection systems.
The strength of our model lies in the integration of CNNs and four-head attention mechanisms for EEG signal analysis. The wavelet transform carefully refines the data points to preserve the main features, facilitating feature extraction by the CNNs.

The incorporation of multi-head attention mechanisms facilitates the model's capacity to attend to diverse temporal segments of the signal simultaneously, enabling the identification of salient interconnections within the EEG data points. Through the assignment of learned attention weights, the model quantifies the relative significance of these temporal relationships in the analytical process. This synergistic integration of methodologies enables a more sophisticated and granular interpretation of EEG signals, potentially enhancing the detection of subtle electrophysiological patterns characteristic of epileptiform activity.
Future research directions encompass comprehensive validation across heterogeneous datasets to evaluate the model's generalizability, alongside focused efforts to augment its robustness and reliability for potential clinical deployment.

%\section*{Declarations}

%\subsection*{Funding}
%No funding was received for conducting this research.

%\subsection*{Conflict of interest/Competing interests}
%The authors declare that there is no conflict of interest. The authors declare that they have no known competing financial interests or personal relationships that could have appeared to influence the work reported in this paper.

%\subsection*{Ethics approval and consent to participate}
%This article does not contain any studies with human participants or animals performed by any of the authors.

%\subsection*{Data availability}
%All data used in this study are available from public repositories. 

%\subsection*{Author contribution}
%This work was carried out in collaboration among all authors. All Authors designed the study, performed the statistical analysis, and wrote the protocol. Authors MG, RM, and AM managed the analyses of the study, managed the literature searches, and wrote the first draft of the manuscript. All authors read and approved the final manuscript.

\bibliography{article}

\end{document}